\documentclass[submission,copyright,creativecommons]{eptcs}
\providecommand{\event}{Special Session: Woman in Login Programming ICLP 2019} 

\usepackage{breakurl}             
\usepackage{underscore}           
\usepackage{amsmath}
\usepackage{amssymb}

\newtheorem{definition}{Definition} [section]

\title{A Temporal Module for Logical Frameworks}

\author{Valentina Pitoni
	\institute{DISIM\\ L'Aquila, Italy}
	\email{valentina.pitoni@graduate.univaq.it}
	\and
 Stefania Costantini
	\institute{DISIM\\ L'Aquila, Italy}
	\email{stefania.costantini@univaq.it}
}
\def\titlerunning{A Temporal Module for Logical Frameworks}
\def\authorrunning{V. Pitoni \& S. Costantini}

\begin{document}
	
	\maketitle
	
	\begin{abstract}
		In artificial intelligence, multi agent systems constitute an interesting
		typology of society modeling, and have in this regard vast
		fields of application, which extend to the human sciences. Logic is
		often used to model such kind of systems as it is easier to verify than other approaches, and provides
		explainability and potential validation. In this paper we define a time module suitable to add time to many logic representations of agents.
	\end{abstract}
	
	\section{Introduction}
	In the literature there different kind of timed logical frameworks exist, where time is specified directly using hybrid logics (cf., e.g., \cite{Hylogic}), temporal epistemic logic (cf., e.g., \cite{TEL}) or simply by using Linear Temporal Logic.	We propose a temporal module which can be adopted to ``temporalize" many logical framework. This module is in practice a particular kind of function that assigns a ``timing" to atoms.
	
	We have exploited this T function in two different settings. The first one is the formalization of the reasoning on the formation of beliefs and the interaction with background knowledge in non-omniscient agents' memory.
	Memory in an agent system is in fact a process of reasoning: in particular, it is the learning process of strengthening a concept. 
	In fact, 
	through memory an agent is potentially able to recall and 
	to learn from experiences so that its beliefs and its future
	course of action are grounded in these experiences. Most of the methods to design agent memorization mechanisms have been inspired by models of human memory \cite{pearson2003effects,logie1995visuo} developed in cognitive science.
	In computational logic, \cite{LoriniBD16} introduces DLEK (Dynamic Logic of Explicit beliefs and Knowledge) as a logical formalization of SOAR (State Operator And Result) Architecture \cite{LairdLR17}, which is one of the most popular cognitive architectures. The underlying idea is to represent reasoning about the formation of beliefs through perception and inference in non-omniscient resource-bounded agents.
	They consider perception, short-term memory (also called ``working memory''), long-term memory (also called ``background knowledge'') and their interaction.
	DLEK is a logic that consists of a static part called LEK (Logic of Explicit beliefs and Knowledge), which is an epistemic logic, and a dynamic component, which extends the static one with ``mental operations''.
	Resource-boundedness in DLEK is modeled via the assumption that beliefs are kept in the short-term memory, while implications that allow reasoning to be performed 
	are kept in the long-term memory.
	New beliefs can be formed in DLEK either from perception, or from previous beliefs in short-term memories and rules in the background knowledge.
	Inferences that add new beliefs are performed one step at a time via an interaction between short- and long-term memories in consequence of an explicit ``mental operation'' that will occur whenever an agent deems it necessary and can allot the needed time \cite{Alechina04,PerlisGKP00}.
	
	The second setting is a logical framework for reasoning about agents’ cognitive attitudes; many formal logics have been proposed for reasoning about concepts taken from qualitative decision theory. Lorini in \cite{Lorini19} proposes a general logical framework for reasoning about agents’ cognitive attitudes of both epistemic type and motivational type. 
	
	The paper is organized as follows. In Section \ref{sect1} we introduce the basic very simple definition of the time function T.
	In Sections \ref{sect2}-\ref{sect3} we discuss the introduction of T in the logics mentioned above.
	In Section \ref{concl} we propose a brief discussion on complexity and on future work and conclude.
	
	\section{The time function T}
	\label{sect1}
	
	In this section we introduce the ``time'' function $T$ that associates to each formula the time interval in which this formula is true.
	To this aim, we assume that each atom has two arguments representing time instants.
	For the sake of simplicity, as we concentrate on these arguments, we ignore all the other arguments;
	i.e., we assume each atom to be of the form $p(t_1,t_2)$.
	\begin{itemize}
		\item $T(p(t_1,t_2))= [t_1,t_2]$, which stands for \textit{``p is true in the time interval $[t_1,t_2]$"} where $t_1,t_2 \in \mathbb{N}$; as a special case we have $T(p(t_1,t_1))= t_1$,  which stands for \textit{``p is true in the time instant $t_1$"} where $t_1 \in \mathbb{N}$ (time instant);
		\item $T(\neg p(t_1,t_2))= [t_1,t_2]$, which stands for \textit{``p is not true in the time interval $[t_1,t_2]$"} where $t_1,t_2 \in \mathbb{N}$;
		\item $T(\varphi \mbox{ op } \psi)= T(\varphi) \biguplus T(\psi)$ with ${op}\in\{\vee,\wedge,\rightarrow\}$, which means the unique smallest interval including both $T(\varphi)$ and $T(\psi)$.
	\end{itemize}
	
	This basic definition, although simple, is able to incorporate a concept of time in virtually any logical formalism
	by creating a link between syntax ans semantics,
	as we show below via two relevant examples.
	Naturally, the T function must then be customized to accommodate the operations which are proper of the 'host' formalism.

	\section{Time Logic of Explicit Belief and Knowledge and Dynamic Logic of Explicit beliefs and Knowledge}
	\label{sect2}
	As in \cite{LoriniBD16}, our logic consists of two different components: a static component, called T-LEK, which is a mix between an Epistemic Logic and Metric Temporal Logic (\cite{MTL90}), and a dynamic component, called T-DLEK, which extends the static one with mental operations, which are very important for ``controlling'' beliefs (addition of belief, update of existing belief, etc.).

	\subsection{T-LEK and T-DLEK Syntax}
	In our scenario we fix $Atm=\{p(t_1,t_2),q(t_3,t_4) \mbox{, ... ,} h(t_i,t_j) \}$ where $t_i \leqslant t_j $ and $p,q,h$ are predicates, that can be equal or not. 
	Moreover $p(t_1,t_2)$ stands for \textit{``p is true from the time instant $t_1$ to $t_2$"} with $t_1, t_2 \in \mathbb{N}$ (\textit{Temporal Representation} of the external world); 
	as a special case we can have $p(t_1,t_1)$ which stands for \textit{``p is true in the time instant $t_1$"}. Obviously we can have predicates with more terms tha only two but in that case we fix that the first two must be those that identify the time duration of the belief (i.e. $ open(1,3,\mbox{door})$ which means ``the agent knows that the door is open from time one to time 3''). Let also $Agt$ be a finite set of agents. 
	
	Below is the definition of the formulas of the language ${\mathcal L}_{T\mbox{-}LEK}$ with $i \in Agt$:
	\begin{center}
		$\varphi, \psi               ~:=~   p(t_1,t_2)\ |\ \neg \varphi\ |\ \square_I\, \varphi\ |\ B_i\,\varphi\ |\ K_i\,\varphi |\ \varphi\ \wedge\ \psi  |\ \varphi\ \rightarrow\ \psi$
	\end{center}

	Others Boolean connectives $\top$, $\bot$, $\leftrightarrow$ are defined from $\neg$ and  $\wedge$ as usual.
	In the formula $\square_{I}\, \Phi$ the MTL Interval ``always'' operator is applied to a formula.
	The operator $B_i$ is intended to denote belief and the operator $K_i$ to denote knowledge. 
	More precisely $B_i$ identifies beliefs present in the working memory, instead $K_i$ identifies what rules present in the background knowledge.
	The language ${\mathcal L}_{T\mbox{-}DLEK}$ of Temporalized DLEK (T-DLEK) is obtained
	by augmenting ${\mathcal L}_{T{-}LEK}$ with the expression $[\alpha]\,\psi$, where $\alpha$ denotes a \emph{mental operation} and 
	$\psi$ is a ground formula.
	The mental operations that we consider are essentially the same as in \cite{LoriniBD16}:
	\begin{itemize}
		\item
		$+ \varphi$, where $\varphi$ is a ground formula of the form $p(t_1,t_2)$ or $\neg p(t_1,t_2)$: the mental operation that serves to form a new belief from a perception~$\varphi$.
		\item
		$\cap(\varphi,\psi)$: believing both $\varphi$ and $\psi$, an agent starts believing their conjunction.
		
		\item
		${\vdash}(\varphi,\psi)$, where $\psi$ is a ground atom, say~$p(t_1,t_2)$:
		an agent, believing that $\varphi$ is true and having in its long-term memory
		that $\varphi$ implies $\psi$ (in some suitable time interval including~$[t_1,t_2]$),
		starts believing that $p(t_1,t_2)$ is true.
		
		\item
		${\dashv}(\varphi,\psi)$  where $\varphi$ and $\psi$ are ground atoms, say $p(t_1,t_2)$ and $q(t_3,t_4)$ respectively:
		an agent, believing $p(t_1,t_2)$ and having in the long-term memory
		that $p(t_1,t_2)$ implies $\neg q(t_3,t_4)$, removes the timed belief $q(t_3,t_4)$
		if the intervals match. Notice that, should $q$ be believed in a wider interval I such that $[t_1,t_2] \subseteq I$, the belief $q(.,.)$ is removed concerning intervals $[t_1,t_2]$ and $[t_3,t_4]$, but it is left for the remaining sub-intervals (so, it is ``restructured'').
	\end{itemize}
	
	\subsection{T-LEK and T-DLEK Semantics}
	Semantics of DLEK and T-DLEK are both based on a set $W$ of worlds.
	In both DLEK and T-DLEK we have the valuation function: $V : W \rightarrow 2^{\mathit{Atm}}$. Also, extend the definition of the ``time'' function $T$:
	\begin{itemize}
		\item $T(B_i \varphi)= T(\varphi)$;
		\item $T(K_i \varphi)= T(\varphi)$;
		\item $T(\square_{I} \varphi)= I$ where $I$ is a time interval in $\mathbb{N}$;
		\item $T([\alpha]\varphi)$ there are different cases depends on which kind of mental operations we applied: 
		\begin{enumerate}
			\item $T(+ \varphi)=T(\varphi)$;
			\item $T(\cap(\varphi,\psi)) = T(\varphi) \biguplus T(\psi)$;
			\item $T({\vdash}(\varphi,\psi))= T(\psi) $;
			\item $T({\dashv}(\varphi,\psi))$ returns the restored interval where $\psi$ is true.
		\end{enumerate}
	\end{itemize}
	For a world $w$, let $t_1$ the minimum time instant of $T(p(t_1,t_1))$ where $p(t_1,t_1) \in V(w)$  and let $t_2$ be the supremum time instant (we can have $t_2 =\infty$) among the atoms in~$V(w)$. Then, whenever useful, we denote $w$ as $w_I$ where $I = [t_1,t_2]$, which
	identifies the world in a given interval.
	
	The notion of LEK/T-LEK model does not consider mental operations, discussed later, and is introduced by the following definition.

	\begin{definition}
		A \emph{ T-LEK model} is a tuple $M=\langle W; N_i; R_i; V; T \rangle$ with $i \in Agt$ where:
		\begin{itemize}
			\item $W$ is the set of worlds;
			\item $V : W \rightarrow 2^{\mathit{Atm}} $ is the valuation function;
			\item $T$ is the ``time'' function;
			\item $R_i \subseteq W{\times}W$ is the accessibility relation, required to be an equivalence relation so as to model omniscience in the background knowledge s.t. $R_i(w)=\lbrace v \in W \mid w_I R_i\ v_{I}\rbrace$ called \emph{epistemic state of the agent $i$ in $w_{I}$}, which indicates all the situations that the agent $i$ considers possible in the world $w_I$ or, equivalently any situation the agent $i$ can retrieve from long-term memory based on what it knows in world $w_I$;
			\item $N_i : W \rightarrow 2^{2^W}$ is a \emph{``neighbourhood''} function,  $ \forall w_I \in W$, $N(i,w_I)$ defines, in terms of sets of worlds, what the agent $i$ is allowed to explicitly \emph{believe} in the world $w_I$;  $ \forall w_{I}, v_{I} \in W$, and $X \subseteq W$:
			\begin{enumerate}
				\item
				if $ X \in N(i,w_I)$, then $X \subseteq R_i(w_{I})$: each element of the neighbourhood is a set composed of reachable worlds;
				\item
				if $w_{I} R_i\ v_{I}$, then $N(i,w_I) \subseteq N(i,v_{I})$: if the world $v_{I}$ is compliant with the epistemic state of world $w_I$, then the agent $i$ in the world $w_I$ should have a subset of beliefs of the world $v_{I}$.
			\end{enumerate}
		\end{itemize}
	\end{definition}
	
	A preliminary definition before the Truth conditions : let $M = \langle W; N_i; R_i; V; T \rangle$ a T-LEK model. Given a formula $\varphi$, for every  $w_I \in W$,\,
	we define $$\parallel \varphi \parallel^M_{w_I} = \{v_I \in W \mid M,v_I \models \varphi\} \cap R_i(w_I).$$
	Truth conditions for T-DLEK formulas are defined inductively as follows:
	\begin{itemize}
		\item $M, w_I \models p(t_1,t_2)$ iff $p(t_1,t_2) \in V(w_I)$ and $T(p(t_1,t_2)) \subseteq I$;
		\item $M, w_I \models \neg \varphi $ iff $M, w_I \nvDash \varphi $ and $T(\neg \varphi)\subseteq I$;
		\item $M, w_I \models \varphi \wedge \psi $ iff $M, w_I \models \varphi $ and $M, w_I \models \psi $ with $T(\varphi),T(\psi) \subseteq I$;
		\item $M, w_I \models \varphi \vee \psi $ iff $M, w_I \models \varphi $ or $M, w_I \models \psi $ with $T(\varphi),T(\psi) \subseteq I$;
		\item $M, w_I \models \varphi \rightarrow \psi $ iff $M, w_I \nvDash \varphi $ or $M, w_I \models \psi $ with $T(\varphi),T(\psi) \subseteq I$;
		\item $M, w_I \models  B_i\, \varphi$ iff \,$\parallel\varphi\parallel^M_{w_I} \in N(w_I)$ and $T(\varphi) \subseteq I$;
		\item $M, w_I \models K_i\, \varphi$ iff for all $v_{I} \in R_i(w_I)$, it holds that $M, v_{I} \models \varphi$ and $T(\varphi) \subseteq I$;
		\item  $M, w_I \models \square_{J} \varphi$  iff $T(\varphi) \subseteq J \subseteq I$ and  for all $v_{I} \in R_i(w_I)$, it holds that  $M, v_{I} \models \varphi$;
	\end{itemize}
	
	Concerning a mental operation $\alpha$ performed by any agent~$i$, we have: 
	$M, w_I \models [\alpha]\,\varphi$ iff $M^{\alpha}, w_I \models \varphi$ and $T(\varphi) \subseteq I$ where
	$M^{\alpha} = \langle W; N^{\alpha}(i,w_I); R_i ; V; T \rangle$.
	Here  $\alpha$ represents a mental operation affecting the sets of beliefs. 
	In particular, such operation can add new beliefs by direct perception, by means of one inference step,  or as a conjunction of previous beliefs.
	When introducing new beliefs, the neighbourhood must be extended accordingly, as seen below; in particular,
	the new neighbourhood $N^{\alpha}(i,                                                                                                                         w_I)$ is defined for each of the mental operations as follows.
	\begin{itemize}
		
		\item Learning perceived belief:
		
		$N^{+ \varphi}(i,w_I) = N(i,w_I)\, \cup \big\{\parallel \varphi \parallel_{w_I}^M\big\}$ with $T(\varphi) \subseteq I$.
		
		The agent $i$ adds to its beliefs perception $\varphi $ (namely, an atom or the negation of an atom) perceived at a time in $T(\varphi)$;
		the neighbourhood is expanded to as to include the set composed of all the reachable worlds which entail $\varphi$ in~$M$.
		
		\item Beliefs conjunction:
		
		$N^{\cap(\psi,\chi)}(i,w_I)=
		\left\{\begin{array}{ll}
		N(i,w_I) \,\cup \big\{\parallel \psi \wedge \chi \parallel_{w_I}^M\big\} & \mbox{if } M,{w_I} \models B_i(\psi) \wedge B_i(\chi) \\ 
		&  \mbox{and } T(\cap(\psi,\chi)) \subseteq I \\
		N(i,w_I)  & \mbox{otherwise}
		\end{array}\right.$
		
		The agent $i$ adds $\psi \wedge \chi$ as a belief if it has among its previous beliefs
		both $\psi$ and $\chi$, with $I$ including all time instants referred to by them;
		otherwise the set of beliefs remain unchanged.
		The neighbourhood is expanded, if the operation succeeds, with those sets of reachable worlds where both formulas are entailed in~$M$.
		
		\item Belief inference:
		
		$N^{\vdash(\psi,\chi)}(i,w_I)= 
		\left\{\begin{array}{ll}
		N(w_I) \,\cup \big\{\parallel \chi \parallel_{w_I}^M\big\} & \mbox{if } M,{w_I} \models B_i(\psi)\ \wedge\ K_i(\psi\rightarrow\chi) \\ 
		&  \mbox{and } T(\vdash(\psi,\chi)) \subseteq I \\
		N(w_I) & \mbox{otherwise}
		\end{array}\right.$
		
		The agent $i$ adds the ground atom $\chi$ as a belief in its short-term memory if it has $\psi$ among its previous beliefs
		and has in its background knowledge $K_i(\psi \rightarrow \chi)$,
		where all the time stamps occurring in $\psi$  and in $\chi$ belong to~$I$.
		Observe that, if~$I$  does not include all time instants involved in the formulas, 
		the operation does not succeed and thus the set of beliefs remains unchanged.
		If the operation succeeds then the neighbourhood is modified by adding $\chi$ as a new belief.

		\item Beliefs revision (applied only on ground atoms).\\ Given $Q=q(j,k)$ s.t. $T(q(j,k))=$ $T(q(t_1,t_2)) \cap T(q(t_3,t_4))$ with $j,k \in \mathbb{N}$ and\\ $P=\big\{$$ M,{w_I} \models  B_i(p(t_1,t_2)) \wedge B_i(q(t_3,t_4)) \wedge K_i(p(t_1,t_2) \rightarrow \neg q(t_3,t_4))$ and $T(\dashv(p(t_1,t_2),$\\$q(t_3,t_4))) \subseteq I$ and there is no interval $J \supsetneq T(p(t_1,t_2))$ s.t. $B_i(q(t_5,t_6))$ where \\ $T(q(t_5,t_6)){=}J\big\}$:
		
		$N^{\dashv(p(t_1,t_2),q(t_3,t_4))}(i,w_I)=
		\left\{\begin{array}{ll}
		N(i,w_I) \setminus \big\{\parallel Q \parallel_{w_I}^M\big\} & \mbox{if P}\\
		N(i,w_I)  & \mbox{otherwise}
		\end{array}\right.$

		The agent $i$ believes that $q(t_3,t_4)$ holds only in the interval $T(q(t_3,t_4))$ and has the perception of $p(t_1,t_2)$ where $T(p(t_1,t_2))\subseteq T(q(t_3,t_4))$. 
		Then, the agent $i$ replaces previous belief $q(t_3,t_4)$ in the short-term memory with $q(t_5,t_6)$ where $T(q(t_5,t_6)){=}T(q(t_3,t_4)) \setminus T(q(t_1,t_2))$. 
		In general, the set $T(q(t_3,t_4)) \setminus T(q(t_1,t_2))$ is not necessarily an interval:
		being $T(p(t_1,t_2))\subseteq T(q(t_3,t_4))$, with $T(p(t_1,t_2)){=}[t_1, t_2]$, and $T(q(t_3,t_4)){=}[t_3,t_4]$, we have that $T(q(t_3,t_4)) \setminus T(q(t_1,t_2)){=} [t_3,t_1-1]{\cup}[t_2+1,t_4]$.
		Thus, $q(t_3,t_4)$ is replaced by $q(t_3,t_1-1)$ and $q(t_2+1,t_4)$ (and similarly if $t_4 = \infty$).
		
	\end{itemize}

	\section{Temporal Dynamic Logic of Cognitive Attitudes}
	\label{sect3}
	The Temporal Dynamic Logic of Cognitive Attitudes (T-DLCA) is an extension of Dynamic Logic of Cognitive Attitudes (DLCA), presented in \cite{Lorini19}. In our extension we introduce the concept of time through a particular function that assigns a ``timing" to knowledge, belief, strong belief, conditional belief, desire, strong desire, comparative desirability and choice.
	
	\subsection{T-DLCA Syntax}
	In our scenario we fix $Atm=\{p(t_1,t_2),q(t_3,t_4) \mbox{, ... ,} h(t_i,t_j) \}$ as in T-LEK framework. 
	Let $Nom=\{x(t_1),$ \\$ y(t_2),  \mbox{, ... ,} z(t_j)\}$ where $x,y,z$ are nominals, which name individual states in models, where $t_j$ are time instants where $j \in \mathbb{N}$. Moreover $Nom$ is disjoint from $Atm$ and let $Agt$ be a finite set of agent.
	
	Below is the definition the language ${\mathcal L}_{TDLCA}$ where $i \in Agt$, $p(t_1,t_2) \in Atm$ and $x(t) \in Nom$:
	\[\begin{array}{rcl}
	\pi, \lambda  & ~:=~ & \equiv_{i}\ |\ \preceq_{i,P}\ |\ \preceq_{i,D}\ |\ {\preceq^{\sim}}_{i,P} \ |\ {\preceq^{\sim}}_{i,D} \ |\ \pi;\lambda \ |\ \pi \cup \lambda \ |\ \pi \cap \lambda \ |\ - \pi \ |\ \varphi? \\ 
	\varphi, \psi              & ~:=~ &  p(t_1,t_2)\ |\ x(t) |\ \neg \varphi \ |\ \varphi \wedge \psi \ |\ [\pi]\varphi \\
	\end{array}\]
	Others Boolean connectives $\top$, $\bot$, $\leftrightarrow$ are defined from $\neg$ and  $\wedge$ as usual. Moreover $\pi$ represents programs which is the basic construct of Dynamic Logic, we can called them \textit{Cognitive Programs} to underline that we are working on reasoning about agents cognitive attitudes; in fact $\pi$ corresponds to a particular configuration of agents cognitive states. As in DL $\pi;\lambda$ stands for ``do $\pi$ followed by $\lambda$'', $\pi \cup \lambda$ stands for ``do  $\pi$ or $\lambda$'', $\pi \cap \lambda$ stands for ``do $\pi$ and $\lambda$'', $- \pi$ is the inverse, $\varphi?$ stands for ``proceed if $\varphi$ is true, else fail''; instead $\equiv_{i}$, $\preceq_{i,P}$, $\preceq_{i,D}$ describe agents knowledge, plausibility and desirability respectively, and ${\preceq^{\sim}}_{i,P}$, $ {\preceq^{\sim}}_{i,D}$ are the complements of $\preceq_{i,P}$, $\preceq_{i,D}$. Also the formula $[\pi]\varphi$ has to be read ``$\varphi $ is true, according to the program $\pi$''; obviously we have different meanings based on the $\pi$ we choose, first of all $[\equiv_{i}]\varphi$ which stands for ``$\varphi$ is true according to what agent $i$ knows'', $[{\preceq^{\sim}}_{i,P}]\varphi$ which stands for ``$\varphi$ is true at all states that, according to agent $i$, are at least as plausible as the current one'' while $[{\preceq^{\sim}}_{i,P}]\varphi$ stands for ``$\varphi$ is true at all states that, according to agent $i$, are not at least as plausible as the current one''. For $\preceq_{i,D}$ and ${\preceq^{\sim}}_{i,D}$ it is enough to replace plausible with desirable in the definition. 
	
	\subsection{T-DLCA Semantic}
	Semantic of T-DLCA is based on a set $W$ of worlds, we have the valuation function: $V : W \rightarrow 2^{\mathit{Atm} \cup \mathit{Nom}}$. We extend the ``time'' function $T$ as follows:
	\begin{itemize}
		\item $T(x(t))= t $
		\item $T([\pi]\varphi)= T(\varphi)$.
	\end{itemize}
	
	\begin{definition}
		A \emph{T-DLCA model} is a tuple $M=\langle W; (\preceq_{i,P})_{i \in Agt}; (\preceq_{i,D})_{i \in Agt};  (\equiv_{i})_{i \in Agt};$\\$ R; V; T \rangle$ where:
		\begin{itemize}
			\item $W$ is the set of worlds defined as in the previous setting;
			\item for every $i \in Agt$, $\preceq_{i,P}$ and $\preceq_{i,D}$ are preorders on $W$ and $\equiv_{i}$ is an equivalence relation on W such that for all $ \tau \in {P,D} $ and for all $w_I, v_J \in W$:
			\begin{enumerate}
				\item $\preceq_{i,\tau} \subseteq \equiv_{i}$ which means that an agent can only compare the plausibility or the desirability of two states, this states have to be in the same interval I;
				\item if $w_I \equiv_{i} v_J$ then $ w_I \preceq_{i,\tau} v_J $ or $w_I \preceq_{i,\tau} v_J$ with $I = J$ which means that the plausibility or the desirability of two states are always comparable if this states are in the same interval I;
			\end{enumerate}
			\item $V : W \rightarrow 2^{\mathit{Atm} \cup \mathit{Nom}} $ is the valuation function and for all  $w_I, v_J \in W$ and $V_{Nom}(w_I)=Nom \cap V(w_I)$:
			\begin{enumerate}
				\item $V_{Nom}(w_I) \neq \varnothing$;
				\item if $V_{Nom}(w_I) \cup V_{Nom}(v_J) \neq \varnothing$ then $w_I = v_J$ with $I=J$;
			\end{enumerate}
			\item $T$ is the ``time'' function;
			\item $R_{\pi} \subseteq W{\times}W$ is a binary relation which works in the following way based on $\pi$:
			\begin{enumerate}
				\item $w_I R_{\equiv_{i}} v_I$ iff $w_I \equiv_{i} v_I$;
				\item $w_I R_{\preceq_{i,\tau}} v_I$ iff $w_I \preceq_{i,\tau} v_I$;
				\item $w_i {\preceq^{\sim}}_{i,\tau} v_I$ iff $w_I \equiv_{i} v_I$ and $w_I \npreceq_{i,\tau} v_I$;
				\item $w_I R_{\pi;\lambda} v_I$ iff $\mbox{ } \exists z_I \in W: w_I R_{\pi} z_I$ and $z_I R_{\lambda} v_I$;
				\item $w_I R_{\pi \cap \lambda} v_I$ iff $w_I R_{\pi} v_I$ or $w_I R_{\lambda} v_I$;
				\item $w_I R_{\pi \cup \lambda} v_I$ iff $w_I R_{\pi} v_I$ and $w_I R_{\lambda} v_I$;
				\item $w_I R_{-\pi} v_I$ iff $v_I R_{\pi} w_I$.
			\end{enumerate}
		\end{itemize}
		The properties ot the valuation function capture the basic properties of nominals: 
		the uniqueness in associating a single nominal with a state. 
	\end{definition}
	Truth conditions for T-DLCA formulas are defined inductively as follows:
	\begin{itemize}
		\item $M, w_I \models p(t_1,t_2)$ iff $p(t_1,t_2) \in V(w_I)$ and $T(p(t_1,t_2)) \subseteq I$;
		\item $M, w_I \models x(t)$ iff $x(t) \in V(w_I)$ and $T(x(t)) \subseteq I$; 
		\item $M, w_I \models \neg \varphi $ iff $M, w_I \nvDash \varphi $ and $T(\neg \varphi)\subseteq I$;
		\item $M, w_I \models \varphi \wedge \psi $ iff $M, w_I \models \varphi $ and $M, w_I \models \psi $ with $T(\varphi),T(\psi) \subseteq I$;
		\item $M, w_I \models [\pi]\varphi $ iff $\forall v_I \in W :$ if $w_I R_{\pi} v_I$ then $ M, v_I \models \varphi $ with $T([\pi]\varphi) \subseteq I$;
		\item $M, w_I \models [?\varphi]\varphi $ iff $\forall v_I \in W :$ if $w_I R_{?\varphi} v_I$ then $ M, v_I \models \varphi $ with $T([?\varphi]\varphi) \subseteq I$ where $w_I R_{?\varphi} v_I$ iff $w_I = v_I$ and $M, w_I \models \varphi$.
	\end{itemize}
	
	\section{Conclusions and Future Work}
	\label{concl}
	In this work we extended two existing approaches (the first one is a logical modeling of short-term and long-term memories in Intelligent Resource-Bounded Agents, the second one is a logical framework for reasoning about agents’ cognitive attitudes of both epistemic type and motivational type) by introducing a time module based on a $T$ function, which manages the time interval when an atom is true. With regard to complexity the $T$ function does not interfere with the 'host' so the overall complexity remains the same for both DLEK and DLCA. Future developments could be the extension to other logical frameworks to prove that we can use this time module in virtually every context.

\nocite{*}	
\bibliographystyle{eptcs}

\end{document}